# Deep Learning-based approaches for automatic detection of shell nouns and evaluation on WikiText-2

## Chengdong Yao[1*], Cuihua Wang[2]

[1]Faculty of Engineering and Information Technology, University of Technology Sydney
[2]School of Foreign Languages, South China University of Technology

**Email address:**

chengdong.yao-1@student.uts.edu.au

[*]Corresponding author

**Abstract:** In some areas, such as Cognitive Linguistics, researchers are still using traditional techniques based on manual rules and patterns. Since the definition of shell noun is rather subjective and there are many exceptions, this time-consuming work had to be done by hand in the past when Deep Learning techniques were not mature enough. With the increasing number of networked languages, these rules are becoming less useful. However, there is a better alternative now. With the development of Deep Learning, pre-trained language models have provided a good technical basis for Natural Language Processing. Automated processes based on Deep Learning approaches are more in line with modern needs. This paper collaborates across borders to propose two Neural Network models for the automatic detection of shell nouns and experiment on the WikiText-2 dataset. The proposed approaches not only allow the entire process to be automated, but the precision has reached 94% even on completely unseen articles, comparable to that of human annotators. This shows that the performance and generalization ability of the model is good enough to be used for research purposes. Many new nouns are found that fit the definition of shell noun very well. All discovered shell nouns as well as pre-trained models and code are available on GitHub.

## 1. Introduction

The concept of the shell noun was first introduced by linguist Schmid, H. J. in 1997 [1]. He defines a shell noun as a functionally defined abstract noun used to convey or refer to complex content in a specific context. Such a noun needs to be identified based on its behavior in a sentence, not based on an inherent meaning. Its actual meaning needs to be determined in context, and because of this, it has an important role to play in the understanding of language. Here are a few examples, with the bold underlined word being the shell noun.

(1) The **fact** that a major label hadn't been at liberty to exploit and repackage the material on CD meant that prices on the vintage LP market were soaring.
(2) The **issue** that this country and Congress must address is how to provide optimal care for all without limiting access for the many.
(3) Living expenses are much lower in rural India than in New York, but this **fact** is not fully captured if prices are converted with currency exchange rates.

In Schmid's study, he suggests 670 words that can be used as shell nouns, while also pointing out that giving an exhaustive list is not possible, as more shell nouns can be found in the right context [1].

There are already some previous researches in attempting to automate the shell noun detection process. Kolhatkar, V. proposed a syntax-based approach to doing Pattern Recognition and achieved accuracies of 62% to 83% [2]. However, their approach is based on sentence structure rather than semantics and function and does not identify shell nouns better in poorly structured sentences. Also, as the hand-written patterns are limited, which is the same limitation as Schmid's, they cannot enumerate all the sentence patterns in which shell nouns occur, so the scope of application is limited [1]. It is also mentioned in their paper that it should be possible to reduce such limitations through a robust Machine Learning approach.

Similar rules and patterns were proposed by Roussel, A. in German [3]. Their work focused more on precision than recall, achieving 56.4% precision but only 26.2% recall. Such accuracy, precision and recall are far from adequate for practical use.

Researchers in Cognitive Linguistics need a more accurate and flexible way of tackling this long-standing task, which is what this paper aims to do. Another task in this paper is to use our proposed model to discover more shell nouns in a more up-to-date dataset and to verify them to test the validity and usefulness of the model.

## 2. Related works

### 2.1. Part-Of-Speech tagging

The task of shell noun detection is like the task of Part-Of-Speech tagging in linguistics. The difference between these two tasks is that Part-Of-Speech tagging focuses more on the overall lexical classification and does not distinguish carefully whether a noun is a shell noun or not. Shell noun detection, on the other hand, focuses on separating shell nouns from other words without regard to lexicality.

The task of Part-Of-Speech tagging has seen more solutions using Deep Learning techniques in addition to the traditional pattern-based approach. In the field of Natural Language Processing, both Part-Of-Speech tagging and Named Entity Recognition can be handled by the same token classification task. For example, Akbik, A., Blythe, D., and Vollgraf, R. used Long Short-Term Memory Neural Network for language sequence modelling and achieved an F1 score of 98.19% on Part-Of-Speech tagging [4]. Their work, therefore, provides a good basis for our research.

### 2.2. Token classification

Shell noun detection is also a Natural Language Processing task. When it comes to Natural Language Processing, an architecture that cannot be avoided is Transformer, a new network architecture based entirely on attention mechanisms, proposed by Vaswani, A. et al. in 2017 [5]. Later Conneau, A. et al. used a dataset of more than 2 TB for pre-training the model, resulting in the XLM-RoBERTa family of models [6].

These models are well suited for language-related tasks, as they take full advantage of the knowledge gained during the pre-training period and require only a small amount of data fine-tuning to achieve good performance. Data for shell noun annotation is very scarce and the subjective nature of the definitions makes data collection more difficult. Fine-tuning a pre-trained model is a better option. Our research will also build on their work.

## 3. Dataset (Corpus)

### 3.1. Annotated shell noun corpus

Data is important for any Deep Learning approach. The annotated shell noun corpus by Simonjetz, F., and Roussel, A. uses a part of the English-German parallel text from Europarl, manually annotated by multiple annotators with rounding [7]. The Europarl corpus is a parallel corpus extracted from the proceedings of the European Parliament by Koehn, P. and was originally intended for the development of Machine Translation system [8].

Due to the subjective nature of the definition of shell noun, the annotated corpus will use to train models directly without annotating them by us. The work in this paper focuses more on English, although the corpus is a parallel corpus, the corresponding German part is ignored. This corpus provides all the data for training, validation and testing. The other dataset mentioned below was used only for the practical application of shell noun detection.

Some statistics for this corpus are shown in Table 1 below.

*Table 1. Statistics for annotated shell noun corpus*

|  | English | German |
|---|---|---|
| Word count | 76,736 | 73,102 |
| Noun count | 18,367 | 14,714 |
| Shell noun count | 1,041 | 1,100 |
| Noun freq. | 23.94% | 20.13% |
| Shell noun freq. | 1.36% | 1.50% |
| Shell proportion | 5.67% | 7.48% |

### 3.2. WikiText-2 dataset

The WikiText-2 dataset is a corpus of verified, excellent and selected articles from Wikipedia by Merity, S., Xiong, C., Bradbury, J., and Socher, R. [9]. The dataset consists of many complete articles, preserving original cases, punctuation and numbers. It is available under CC BY-SA. It was chosen for shell noun detection because it is of moderate size, with about 2.5 million words; and relatively new, the dataset having been published in 2016. This gives a higher probability of finding more eligible shell nouns.

In addition, Wikipedia is a website that many people use, and the text on it should be closer to the text that people can see in their daily lives. The Europarl corpus from the proceedings of the European Parliament, on the other hand, is less relevant to the general public.

The difference in statistics between WikiText-2 and the corpus mentioned in 3.1 is shown in Table 2 below.

*Table 2. Difference between annotated shell noun corpus and WikiText-2*

|  | Annotated | WikiText-2 |
|---|---|---|
| Word count | 76,736 | 2,472,519 |
| Noun count | 18,367 | 749,016 |
| Noun freq. | 23.94% | 30.29% |

## 4. Approaches

### 4.1. Review of the traditional approach

There are two main steps in shell noun detection using the traditional method:

(1) Lexico-grammatical analysis and labelling of the components of each word using relevant software.
(2) Matching against the patterns proposed by Hans Schmid and later researchers [1].

*Table 3. 10 patterns proposed by Schmid, H. J.*

| | |
|---|---|
| Noun-be-to | Noun-wh |
| Noun-be-that | Noun-of |
| Noun-be-wh | th-Noun |
| Noun-to | th-be-Noun |
| Noun-that | Sub-be-Noun |

This approach relies on lexico-grammatical analysis, and the accuracy of identifying shell nouns also relies on the accuracy of component annotation. Nowadays Part-Of-Speech tagging based on Deep Learning is well established and the

Part-Of-Speech tagging model proposed by Akbik, A. et al. is used to reproduce Kolhatkar, V.'s algorithm [2, 4]. Even though the accuracy of Part-Of-Speech tagging is very high, their algorithm still does not perform well on our test set, as can be seen in Chapter 6.2 of this paper.

The reason for this is explained by themselves. Since it is impossible to exhaust all patterns of occurrence of shell nouns, some words fit the definition but not the pattern. And the reason why this approach does not work well is that it relies entirely on lexical and grammatical usage. The role of a shell noun is to convey or refer to a particular piece of information in the context. A better way to address this type of problem requires consideration of the underlying semantics of the words.

### 4.2. Our LSTM-based approach

Long Short-Term Memory (LSTM) Neural Networks are a very practical type of network for sequence data processing [10]. Articles are sequences made up of many words and are therefore also suitable for processing using LSTM. Our proposed LSTM-based model is built using Flair and PyTorch, and the general architecture is shown in Figure 1. The model uses three different sets of word vectors as input, GloVe, Flair-news-forward and Flair-news-backward [11]. They go into an LSTM and linear module, respectively, and are then combined. It is then connected to a larger LSTM module and a linear module. Eventually, it will pass through the Viterbi and the Conditional Random Field to get the final output [12, 13].

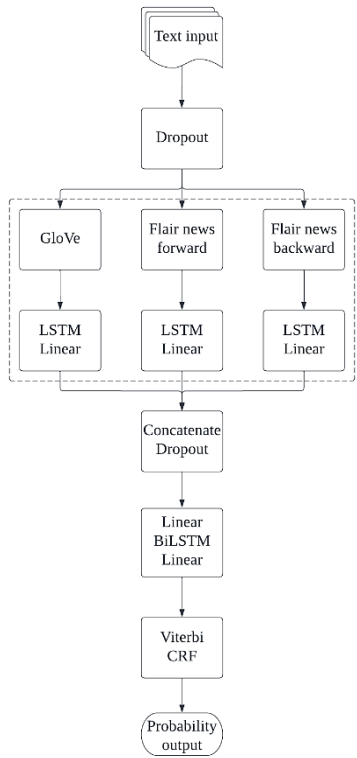

*Figure 1. The general architecture of our LSTM-based model*

Unlike the Transformer-based model below, the LSTM-based model is commonly used to deal with language problems by using the word vector corresponding to each word as input. Three different word vectors are used to give the model more information. GloVe uses co-occurrence matrices to obtain features of words both locally and as a whole, while the word vectors starting with Flair-news obtain features by allowing the language model to predict words forwards and backwards respectively. Doing so also enhances the robustness of the model, as these word vectors are constructed differently and use different training data.

When reasoning on a larger dataset, words will likely appear that have never been seen in the training data. And the scenario of dropping certain words with some probability to let the model infer it from other words is like the model encountering words it has never seen before. So, the interspersed dropout layer acts as a regularization to suppress the overfitting of the model and improve generalization.

As with many other language tasks, the current word alone is not sufficient to infer whether it is a shell noun, so the Viterbi algorithm and Conditional Random Field are used to find more likely correspondences. Both techniques are also frequently used in Automatic Speech Recognition and Natural Language Processing.

### 4.3. Our Transformer-based approach

Since the Transformer architecture was proposed in 2017, it has been successful in areas such as Natural Language Processing and Computer Vision. In our proposed Transformer-based approach, a RoBERTa model is used to replace the word vectors in the LSTM-based approach [14]. All LSTM modules are removed so that the model is informed only by the features extracted by RoBERTa. The model uses the XLM-RoBERTa-base hosted on Hugging Face's Model Hub [15]. Its architecture is shown in Figure 2. As all word vectors and LSTM modules are replaced by a RoBERTa and linear module, this approach is much cleaner in terms of process.

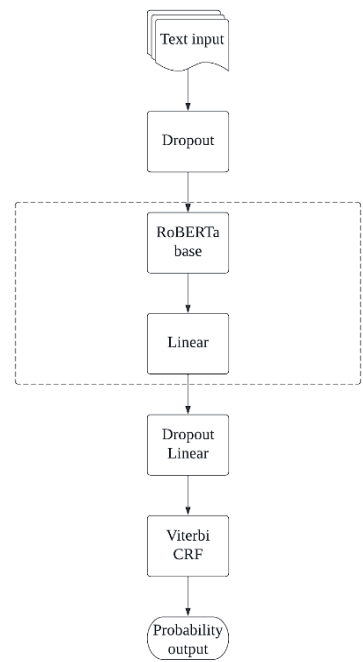

*Figure 2. The general architecture of our Transformer-based model*

One of the advantages of this approach is that the word vectors generated by Transformer are strongly context-dependent, i.e., the word vectors for the same word in different contexts are different. GloVe, for example, has a fixed word vector for each word after training, and therefore cannot be better adapted to changing contexts.

Another advantage is that, due to resource constraints, we cannot provide a huge dataset for the LSTM-based model to be trained. In contrast, the XLM-RoBERTa family of models are pre-trained on a 2 TB dataset. If only fine-tune it at a very low learning rate, the knowledge learned by the model during pre-training will be retained, which is one of the reasons why pre-trained language models are useful. And this may be even more useful for the model to discover shell nouns in passages that have never been seen before.

Since the Transformer-based model has a higher number of parameters than the LSTM-based model, it is less efficient on short inputs. However, when the input sequence becomes longer, the Transformer is more efficient than the LSTM. This is because Transformer's operations are performed in parallel at the sequence level, whereas LSTM is serial. As the WikiText-2 dataset has over 2 million words, Transformer will be more efficient on data of this size, as longer input lengths can be used.

### 4.4. The loss functions

The simplest loss function to solve this task is the cross-entropy loss, which will optimize the prediction loss for each word individually. If the prediction loss for each word is small enough, the model will naturally achieve good performance. In this scenario, what the model learns is the probability of each word being a shell noun in the corresponding context. The focus is on the words themselves. Cross-entropy loss has been applied to several different tasks and its formula will be set out again below for comparison purposes.

$$L = -\frac{1}{N}\sum_{i=1}^{N}\sum_{c=1}^{M} y_{ic} \log(p_{ic})$$
(1)

However, this ignores the information about the regularity between words. So, there is a better way, which is to use a Conditional Random Field, as in the paper by Huang, Z., Xu, W., and Yu, K. [16]. After introducing CRF to the model, it is no longer possible to use cross-entropy as a loss function. This is because the focus when training the model will no longer be on each word itself, but on the transition probabilities between different states. Specifically, it is the transition probability between whether the words of this sentence are shell nouns or not in our task. In the process of forwarding inference, the sequence with the highest overall probability needs to be found, and this can be solved using the Viterbi algorithm. This is the reason why the Viterbi module is used.

As the loss of each word is now no longer optimized separately, but rather the loss of transition probabilities, a new loss function needs to be used. It is generally referred to as the Viterbi loss or CRF loss and here is its formula.

$$L = -\log\left(Softmax\left(\sum_{i=1}^{N} P_T(y_i|y_{i-1}) + \sum_{i=1}^{N} P_E(y_i|p_i)\right)\right)$$
(2)

In our specific task, $P_T(y_i|y_{i-1})$ in Formula (2) above denotes the transition probability of the state of the i[th] word given the state of the (i-1)[th] word. $P_E(y_i|p_i)$ denotes the emission probability of the state of the i[th] word given its word in the corresponding context. Both transition probabilities and emission probabilities are trainable parameters in CRF.

## 5. Ablation study

### 5.1. Is POS lexical information necessary?

In the traditional pattern-based approach, it is necessary to determine whether a word is a shell noun based on the results of lexical annotation, so this is where lexical information is necessary. When using a Deep Learning-based approach, the text itself can be input and the probability that each word is a shell noun can be output directly. In this case, is it still necessary to retain lexical information? Relevant experiments are designed to answer this question.

The experimental results in Chapter 6 show that, both in LSTM and Transformer, when the Viterbi and CRF are used, the inclusion of lexical information will, on the contrary, degrade the performance of the model. However, if the Viterbi and CRF are not used, the addition of lexical information improves the performance of the model. However, the lexical information provides a limited improvement, and the gain is not comparable to that of CRF.

Lexicality, as additional information about the text, could give more supervisory information to help the model achieve better performance when the model itself is not good enough. However, with CRF, the model itself is already good enough. Adding further lexical information would instead distract the model from the single task of whether it is a shell noun or not. Therefore, in our final model, CRF instead of lexical information is used to obtain better performance.

### 5.2. Is the Viterbi module and CRF necessary?

The Viterbi and CRF are used for tasks such as Part-Of-Speech tagging, because the occurrence of words is conditional, also known as the probability of state transfer. Most of the relevant tasks use CRF, but is the Transformer model powerful enough to perform well without using it? Or does the removal of CRF help the model? Experiments are also designed to investigate this question by replacing CRF with a cross-entropy loss, as described in Chapter 4.4.

However, the experimental results in Chapter 6.2 show that in most cases, removing CRF harms the model, especially in the Transformer-based model. A Transformer-based model without CRF performs even worse than an LSTM-based model that also does not use CRF. This may also reflect the fact that although both LSTM and Transformer are good at working with sequential data, their respective strengths are different.

Given this, even a strong Transformer model needs to use traditional tools such as the Viterbi and CRF to achieve better performance when encountering certain problems, such as Part-Of-Speech tagging and shell noun detection. For this reason, these two modules are used in both of our proposed models.

## 6. Results and comparison

### 6.1. Performance on the training set

*Table 4. Performance on the training set*

| # | Model type | Use POS | Use CRF | Precision (%) | Recall (%) | F1 (%) |
|---|---|---|---|---|---|---|
| 1 | LSTM | × | × | 92.80 | 91.53 | 92.16 |
| 2 | LSTM | × | √ | 92.34 | 86.92 | 89.55 |
| 3 | LSTM | √ | × | **93.26** | **92.11** | **92.68** |
| 4 | LSTM | √ | √ | 92.62 | 78.82 | 85.16 |
| 5 | Transformer | × | × | 89.09 | 90.52 | 89.80 |
| 6 | Transformer | × | √ | **99.88** | 99.26 | 99.57 |
| 7 | Transformer | √ | × | 91.18 | 91.51 | 91.34 |
| 8 | Transformer | √ | √ | **99.88** | **99.75** | **99.82** |

In the training set, the use of lexical information brings a boost to performance for both LSTM-based and Transformer-based models. This suggests that it helps to help the model fit better on the training set. Furthermore, using the Viterbi and CRF in the LSTM model leads to performance degradation in the training set, while its use in the Transformer model leads to a boost.

### 6.2. Performance on the testing set

*Table 5. Performance on the testing set*

| # | Model type | Use POS | Use CRF | Precision (%) | Recall (%) | F1 (%) |
|---|---|---|---|---|---|---|
| 1 | LSTM | × | × | 81.69 | **84.06** | 82.86 |
| 2 | LSTM | × | √ | 89.74 | 82.35 | **85.89** |
| 3 | LSTM | √ | × | 86.52 | 83.70 | 85.08 |
| 4 | LSTM | √ | √ | **94.59** | 76.92 | 84.85 |
| 5 | Transformer | × | × | 78.38 | 84.06 | 81.12 |
| 6 | Transformer | × | √ | **93.98** | 82.98 | **88.14** |
| 7 | Transformer | √ | × | 81.08 | **86.96** | 83.92 |
| 8 | Transformer | √ | √ | 88.64 | 86.67 | 87.64 |

The traditional pattern-based approach is reimplemented in the testing set. Even with the very powerful POS tagging model, the recall and precision of the traditional approach are only about 55%. This shows that it is not the accuracy of the lexical annotation that limits the traditional approach but the patterns themselves.

For both of our proposed models, the best versions of each use the Viterbi and CRF, and the comparison of the F1 scores shows that CRF reduces the degree of overfitting of the model and enhances generalization. Lexical information, on the other hand, negatively affects the performance of the model, both in the LSTM-based model and the Transformer-based model. Considering that this information is very useful in the training set, it affects the generalization ability of the model.

### 6.3. Comparison with the traditional approach and human

*Table 6. Performance comparison on the testing set*

| # | Approach | Precision (%) | Recall (%) | F1 (%) |
|---|---|---|---|---|
| 2 | LSTM | 89.74 | 82.35 | 85.89 |
| 6 | Transformer | 93.98 | 82.98 | 88.14 |
| 9 | Patterns | 56.18 | 54.35 | 55.25 |
| 10 | Human | 91.90 | 88.33 | 90.08 |

Language is uncertain, so the words that each annotator considers to be shell nouns are not the same. The confusion matrix for annotators is given in Simonjetz, F. and Roussel, A.'s paper and can be used to work out how well human annotators perform [7]. This is how the data for #10 in Table 6 is calculated.

As can be seen, our proposed LSTM model has a 30% performance improvement over the traditional approach, and our proposed Transformer model has a 33% improvement. This directly improves the F1 score of the model in the testing set from around 55% to around 88%, reaching a practical level.

Affected by a lower recall, the F1 score of the model is below the human level. However, our Transformer-based model's precision on the testing set has narrowly outperformed that of the human annotator. This is a height that no previous work has achieved.

### 6.4. Parameters and efficiency

*Table 7. Parameters and efficiency of proposed models*

|  | LSTM | Transformer |
|---|---|---|
| Parameters | 64M | 278M |
| Efficiency | 700 words/second | 7857 words/second |

Our test platform uses an RTX 3070 Laptop. As can be seen from Table 7 above, our LSTM model is a relatively lightweight model, with only about 64 million parameters, requiring little video memory. However, due to its serial computing nature, it performs poorly in terms of efficiency on long passages, processing only about 700 words per second. The Transformer model has a relatively large number of parameters, 278 million, but this is a small number of parameters compared to common language models. Moreover, due to its parallel computing nature, the processing efficiency on long passages is much better than that of LSTM, exceeding 7800 words per second.

## 7. Evaluation on WikiText-2

### 7.1. Overall statistics

*Table 8. Statistics on the results of our Transformer model on WikiText-2*

| Word count | 2,472,519 |
|---|---|
| Noun count | 749,016 |
| Shell noun count | 21,021 |
| Noun freq. | 30.29% |
| Shell noun freq. | 0.85% |
| Shell proportion | 2.81% |
| All shell nouns | 959 |
| New shell nouns | 522 |

Our proposed Transformer-based model with the highest F1 score is taken to inference on WikiText-2, and the results are shown in Table 8. The model found a total of 21,021 shell nouns in this dataset, representing 0.85% of all nouns. This proportion is lower compared to the training data. However, given that the model has a recall of around 83%, approximately 1.02% of the nouns in this dataset should be shell nouns. This may be because Wikipedia uses more usual simple language, as opposed to the more formal style of the proceedings of the European Parliament.

Our model also found 522 new shell nouns that were not among Schmid, H. J.'s 670 words, but also qualify as shell nouns [1]. The full list of 959 shell nouns in WikiText-2 can be found on our GitHub repository. Below are 10 selected new words with the highest frequency of occurrence for manual verification.

*7.2. Manual verification*

(1) Name
… quell a tribal incursion into Pontus; he gives the **name** of assassin as *another Odaenathus who may or may not* …

(2) Review
… target. Sal Cinque Mani of slant magazine gave a negative **review** to the video, describing it as "*sloppily edited*…

(3) Title
… minutes in a cafe and came up with the **title** of *stop!! Habari @-@ Kun!* as a reference to …

(4) Route
… North Dakota. Later that year, Bourgmont published the route to *be taken to the Missouri river*, the first …

(5) Work
… of them. At the same time, our main **work** is to *improve the environment, make people more able and* …

(6) Record
… ever, in *four minutes and 33 seconds*. His **record** stood for twenty years until broken by Sadio Mané on 16 …

(7) Reference
… as the primary interface. Brown had considered updating the **reference** to advertising Star Wars: *the force unleashed because loom was* …

(8) Circumstance
… in allowing Gwen to progress with her wedding in the **circumstance** of *her being pregnant*. She felt that the episode had …

(9) Nomination
… the workhouse donkeys. In 1963 Olivier received another bafta **nomination** for *his leading role as a schoolteacher accused of sexually molesting* …

(10) Success
… 1206. The Song had depended on Wu's **success** in the west to *divert Jin soldiers away from the eastern* …

Above are 10 examples of the most frequently occurring shell nouns newly discovered by our proposed model, all samples were randomly extracted from the WikiText-2 dataset. Words underlined in bold are shell nouns, while those in italics are the part of the message they refer to or convey, i.e., the shell content. Manual verification can be used as a gold standard. The 10 randomly selected examples all meet the functional definition of shell noun, and all refer to or convey context-specific information. This means that the model does have the ability to discover new shell nouns in texts that have never been seen before, and the performance is as expected.

## 8. Conclusion

This paper investigates the practical application of two classical neural network architectures, Long Short-Term Memory and Transformer, to the task of shell noun detection in the traditional field of Cognitive Linguistics. A new model based on LSTM, and another based on Transformer are proposed. Its performance is improved by 30% to 33% over the traditional pattern-based approach, while also automating the entire process. The probability of each word being a shell noun is obtained by simply entering the text. The impact of lexical information, the Viterbi and Conditional Random Fields on this task was also investigated. Shell noun detection was evaluated on the WikiText-2, where a total of 522 new shell nouns were found that met the requirements but had not been previously proposed, and the 10 with the highest frequency were manually validated.

The code for the new model proposed in this paper and the 959 shell nouns with their frequency statistics are publicly available at https://github.com/ycd2016/shellnouns. The trained models have also been uploaded to Google Drive and the link can be found on the GitHub page above. This work not only brings new Deep Learning-based solutions to the field but also brings more researchable content. For example, revisiting models written by previous researchers as well as exploring new words discovered by the models. We look forward to more collaborations across knowledge domains to promote the joint development of different disciplines.